\documentclass[lettersize,journal]{IEEEtran}
\usepackage{amsmath,amsfonts}
\usepackage{algorithmic}
\usepackage{algorithm}
\usepackage{array}
\usepackage[caption=false,font=normalsize,labelfont=sf,textfont=sf]{subfig}
\usepackage{textcomp}
\usepackage{stfloats}
\usepackage{url}
\usepackage{verbatim}
\usepackage{graphicx}
\usepackage{cite}
\usepackage{booktabs}  
\usepackage{multirow}
\usepackage[justification=centering]{caption}
\usepackage[numbers]{natbib}
\usepackage{hyperref}
\hypersetup{
    colorlinks=true,
    linkcolor=blue,  
    citecolor=green,  
    urlcolor=green,   
}
\begin{document}

\title{Advancing Out-of-Distribution Detection through Data Purification and Dynamic Activation Function Design}

\author{Yingrui Ji, Yao Zhu, Zhigang Li, Jiansheng Chen, Yunlong Kong* and Jingbo Chen %~\IEEEmembership{Staff,~IEEE,}
        % <-this % stops a space
\thanks{Yingrui Ji, Jiansheng Chen, Yunlong Kong and Jingbo Chen is with the Aerospace Information Research Institute, Chinese Academy of Sciences, Beijing 100190, China, also with the University of Chinese Academy of Sciences, Beijing 100190, China. }
\thanks{Yao Zhu is with the Zhejiang University, Hangzhou 310027, China.}
\thanks{This work was supported by the National Key R\&D Program of China under Grant 2021YFB3900504. (Corresponding author: Yunlong Kong.)}
}
%\thanks{This paper was produced by the IEEE Publication Technology Group. They are in Piscataway, NJ.}% <-this % stops a space
%\thanks{Manuscript received April 19, 2021; revised August 16, 2021.}}

% The paper headers
\markboth{IEEE Transactions on Circuits and Systems for Video Technology, December~2023}%
{Shell \MakeLowercase{\textit{et al.}}: A Sample Article Using IEEEtran.cls for IEEE Journals}

\IEEEpubid{0000--0000/00\$00.00~\copyright~2023 IEEE}
% Remember, if you use this you must call \IEEEpubidadjcol in the second
% column for its text to clear the IEEEpubid mark.

\maketitle

\begin{abstract}

In the dynamic realms of machine learning and deep learning, the robustness and reliability of models are paramount, especially in critical real-world applications. A fundamental challenge in this sphere is managing Out-of-Distribution (OOD) samples, significantly increasing the risks of model misclassification and uncertainty. Our work addresses this challenge by enhancing the detection and management of OOD samples in neural networks. We introduce OOD-R (Out-of-Distribution-Rectified), a meticulously curated collection of open-source datasets with enhanced noise reduction properties. In-Distribution (ID) noise in existing OOD datasets can lead to inaccurate evaluation of detection algorithms. Recognizing this, OOD-R incorporates noise filtering technologies to refine the datasets, ensuring a more accurate and reliable evaluation of OOD detection algorithms. This approach not only improves the overall quality of data but also aids in better distinguishing between OOD and ID samples, resulting in up to a 2.5\% improvement in model accuracy and a minimum 3.2\% reduction in false positives. Furthermore, we present ActFun, an innovative method that fine-tunes the model's response to diverse inputs, thereby improving the stability of feature extraction and minimizing specificity issues. ActFun addresses the common problem of model overconfidence in OOD detection by strategically reducing the influence of hidden units, which enhances the model's capability to estimate OOD uncertainty more accurately. Implementing ActFun in the OOD-R dataset has led to significant performance enhancements, including an 18.42\% increase in AUROC of the GradNorm method and a 16.93\% decrease in FPR95 of the Energy method. Overall, our research not only advances the methodologies in OOD detection but also emphasizes the importance of dataset integrity for accurate algorithm evaluation. By refining the distinction between in-distribution and out-of-distribution data, our contributions aim to enhance the model's proficiency in identifying and generalizing from unknown data, thereby ensuring greater model reliability in diverse applications.
%We will open-source the datasets combination and code once the paper is accepted. 
\end{abstract}

\begin{IEEEkeywords}
Out-of-Distribution detection, OOD datasets, In-Distribution datasets, OOD evaluation.
\end{IEEEkeywords}

\section{Introduction}
\IEEEPARstart{T}{he} increasing significance of Out-of-Distribution (OOD) detection in deep neural networks is underscored by its crucial role in enhancing network security and reliability\cite{hendrycks2016baseline, liu2020energy, yang2021generalized, yang2022openood}. Despite their impressive capabilities, deep neural networks can produce unreliable predictions when encountering inputs outside their training distribution. This unreliability poses a considerable risk in safety-critical applications, such as medical diagnostics\cite{schlegl2017unsupervised} and autonomous vehicles\cite{kitt2010visual}, where classifier dependability is imperative.

OOD detection is primarily concerned with distinguishing uncertain OOD predictions from more reliable In-Distribution (ID) predictions. The vital role of OOD detection in ensuring the safe deployment of machine learning systems is highlighted, especially in open-world settings\cite{drummond2006open} where input data distributions are inherently unpredictable.
It serves a dual purpose: reducing the likelihood of false predictions and bolstering the model's credibility and practicality in real-world applications. OOD detection hinges on accurately estimating data density or depicting features within a distribution, a task made challenging by the complex nature of data distributions. Typically, models are pre-trained on in-distribution (ID) data, which often covers a limited range, contrasting starkly with the diverse and multifaceted nature of real-world data.  

\begin{figure*}[!t]
\centering
\includegraphics[width=7in]{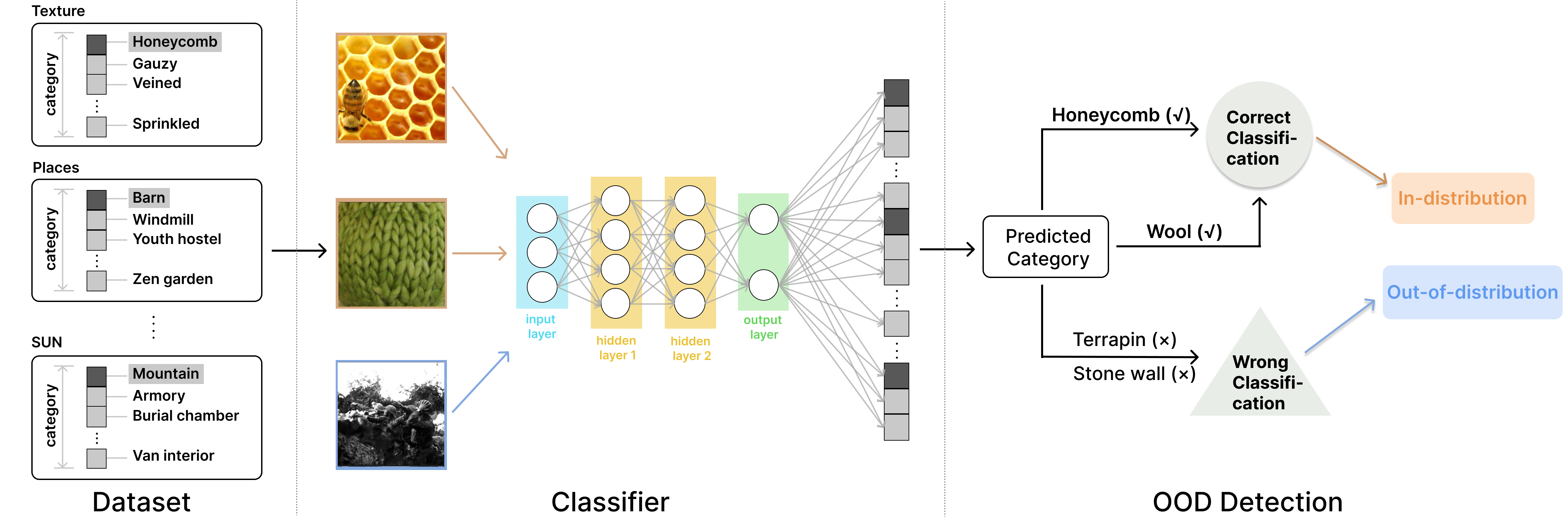}
\caption{The illustration of the out-of-distribution (OOD) detection process adopted by our classifier. Texture\cite{cimpoi2014describing}, Places365\cite{zhou2017places}, SUN subset\cite{xiao2010sun} and other datasets are taken as input.  The classifier predicts the input data through the network category. The rightmost part of the figure highlights the OOD detection mechanism. It shows that some samples (such as the two pictures shown with orange borders) are correctly classified and, therefore, considered IDs, but some samples (such as the blue picture shown in the border) are accurately identified as OOD. This shows that some samples of classification pairs are mistakenly placed in the OOD test set.}
\label{fig_1}
\end{figure*}

In the increasingly scrutinized realm of OOD detection tasks, assessing the performance of various detection algorithms becomes a critical topic, which determines how to make fair and effective comparison. However, we've noticed a crucial issue: the OOD datasets commonly used in the conventional evaluation always contain a substantial number of ID samples as shown in Fig. \ref{fig_1}. The conventional  evaluation methods require detection algorithms to differentiate between the OOD dataset and the ID dataset. Yet, when the OOD dataset includes ID samples (noise data), the expected behavior would be to identify this noise data as ID samples and the rest of the OOD dataset as OOD. However, this approach might yield lower evaluation results because conventional evaluation methods mandate that the detection algorithm categorizes all samples within the OOD dataset as OOD samples.

To address these issues, we have undertaken the crucial task of purifying the OOD dataset. This purification process involves the meticulous removal of mislabeled ID samples, thereby ensuring the integrity and clarity of the OOD dataset. Training models with a purified OOD dataset better equips them to mirror real-world conditions, where the separation between ID and OOD data is not always clear-cut. The use of pre-trained models on such purified datasets is aimed at  enhancing their robustness, aligning with the core goal of OOD detection—to effectively generalize across various environments and reliably identify novel, unseen OOD instances. \newpage\noindent
This approach not only underscores the importance of dataset purity in OOD detection but also highlights our commitment to refining methodologies for more accurate and reliable model performance in practical applications.

After analyzing the potential negative impact of noise within the dataset on OOD detection tasks and constructing a purified dataset, we further investigated methods to enhance the performance of existing OOD detection algorithms, termed ActFun. OOD detection is a single-sample hypothesis testing task, where the detection outcome of an individual sample might be influenced by the sample's specificities, leading to less robustness. Hence, we propose conducting detection within the input's neighborhood. Specifically, instead of calculating the activation of a single input, we compute the expectation within the input's neighborhood. Moreover, we derived a simplified formula for computing this expectation theoretically.
A key benefit of ActFun is its ability to improve the separability between ID and OOD data distributions, resulting in notable enhancements in the area under the receiver operating characteristic curve (AUROC), from 49.35\% to 67.77\%, and a significant reduction in the false positive rate of OOD (negative) examples when the true positive rate of in-distribution (positive) examples is as high as 95\% (FPR95), from 82.6\% to 65.67\%.
%pic 1----ood pic path: \imagenet-o\n02395406

Additionally, our study includes an analysis exploring the underlying mechanisms of ActFun's contribution to OOD detection. We demonstrate ActFun’s effectiveness, especially in scenarios where OOD activations exhibit increased chaos and positive skewness compared to ID activations---a characteristic frequently observed in numerous OOD datasets. A comprehensive evaluation of widely-recognized OOD detection benchmarks confirms ActFun's superior performance relative to established baseline methods.

In summary, our research addresses the pressing issues of dataset noise and the refinement of evaluation techniques in OOD detection, presenting these key contributions as pivotal advancements in the field:

\begin{enumerate}
    \item We present the OOD-R dataset, an innovative amalgamation of existing open-source datasets, distinguished by its low noise level. This rectified dataset, through strategic noise filtering, offers enhanced data quality for OOD detection, providing clearer and more reliable samples for research and model development.
    \item We have also introduced the ActFun activation structure, which substitutes traditional ReLU with the expectation version of ReLU in various networks. This change significantly boosts OOD detection's specificity and accuracy. Notably, ActFun has shown a considerable improvement in evaluation methods, marked by up to 18.42\% increase in AUROC and a minimum of 16.93\% decrease in FPR95, underscoring the importance of precise hyperparameter calibration in optimizing OOD detection.
    \item Our research has examined the impact of the hyperparameter $\beta$ on different OOD detection algorithms. We found a strong correlation between this parameter and each method's performance, highlighting the need for accurate hyperparameter tuning, especially when modifying activation functions, to enhance OOD detection effectiveness.
\end{enumerate}

\section{Related Work}
OOD detection \cite{zhu2022rethinking,zhu2022boosting} plays a pivotal role in ensuring the reliability and robustness of machine learning models, especially in computer vision. Accurately identifying and processing data that significantly deviates from the training distribution is essential in real-world scenarios characterized by unpredictable variations. This section delves into datasets, the core methodologies and key findings within the OOD detection domain, underscoring their contributions and limitations in propelling the field forward.

\textbf{Diverse datasets for OOD model evaluation.} In our research, we utilize ImageNet\cite{deng2009imagenet} as the primary ID dataset, encompassing approximately 14 million images across over 20,000 categories. ImageNet's extensive usage in visual object recognition research makes it a cornerstone dataset for numerous computer vision endeavors. We incorporate five diverse open-source datasets for OOD evaluation, each offering unique challenges and perspectives. These include Texture\cite{cimpoi2014describing}, known for its varied range of natural textures; ImageNet-O\cite{hendrycks2021natural}, a subset of ImageNet curated explicitly for its challenging OOD properties; iNaturalist\cite{van2018inaturalist}, which covers a wide array of biological species; Places365\cite{zhou2017places}, featuring a variety of natural and urban scenes; and the SUN subset\cite{xiao2010sun}, focusing on a broad range of indoor environments. These datasets, distinct from ImageNet categories, are integral in assessing our model's classification capabilities across varied scenarios.

\textbf{Strategies for generating models.} Generative models are a noteworthy strategy in OOD detection, estimating input data's probability density\cite{kingma2013auto, tabak2013family, rezende2014stochastic, dinh2016density, van2016conditional, huang2017stacked}. However, they sometimes misclassify OOD data as high likelihood\cite{nalisnick2018deep} and present challenges in training and optimization, often underperforming compared to discriminative models. Our work, therefore, concentrates on discriminative-based approaches for OOD detection. Despite the theoretical appeal of generative models\cite{kirichenko2020normalizing, ren2019likelihood, schirrmeister2020understanding, serra2019input, wang2020further, xiao2020likelihood}, limitations make them less suitable for large-scale OOD detection goals. People prioritise enhancing the robustness and scalability of methods. Another research direction involves incorporating auxiliary outlier data for model regularization\cite{bevandić2018discriminative, geifman2019selectivenet, malinin2018predictive, mohseni2020self, subramanya2017confidence, liu2020energy}. This includes both realistic\cite{hendrycks2018deep, mohseni2020self, papadopoulos2021outlier, liu2020energy, chen2020robust} and synthetic images generated by GANs\cite{lee2017training}. Our approach diverges by refining the model using only in-distribution data, avoiding the complexities of compiling and integrating external anomaly datasets, and streamlining the model development process while focusing on practical, scalable solutions for effective OOD detection.

\textbf{Development of evaluation methods for OOD detection.}The OOD detection landscape has seen significant advancements over recent years. Nguyen et al.\cite{nguyen2015deep} highlighted deep neural networks' susceptibility to adversarial attacks, introducing methods to assess network reliability using adversarial samples. Hendrycks and Gimpel\cite{hendrycks2016baseline} set a baseline with MSP\cite{hendrycks2016baseline}, leveraging the softmax output's inherent uncertainty for OOD detection. Lee et al.\cite{lee2018simple} improved OOD detection using Mahalanobis distance within the feature space. Liu et al.\cite{liu2020energy}'s energy-based method furthered this progress by utilizing network energy estimations for OOD discernment. The Generalized ODIN method\cite{hsu2020generalized, liang2017enhancing} introduced temperature scaling and peak adjustments for enhanced performance. Recent developments include Wang et al.\cite{wang2022vim}'s approach, combining virtual adversarial training with logical probability matching, and Hendrycks et al.\cite{hendrycks2019scaling}'s KL-Matching method, focusing on probability distribution differences for unknown data evaluation. Sun et al.\cite{sun2021react}'s ReAct model employs post-hoc unit activation modifications, aligning activation patterns with optimal performance scenarios. Our ActFun method, in comparison, facilitates smoother transitions in learning feature representations, thereby enhancing OOD detection. Lin et al.\cite{lin2021mood}'s multi-level feature extraction technique and the Model Output Statistics\cite{huang2021mos} approach have shown promise in OOD detection. However, each has its limitations and strengths, particularly in scalability and effectiveness across varied dataset sizes.

\section{Method}

In the burgeoning era of artificial intelligence (AI), the quality and integrity of datasets have become paramount. As AI models evolve, transcending essential pattern recognition to achieve nuanced understanding and reasoning, the role of datasets, particularly those handling OOD samples, is critical in ensuring model robustness and reliability. OOD datasets, characterized by their variability and noise, mirror the unpredictability and complexity of real-world scenarios. In such an environment\cite{zhu2021rethinking}, cleaning and refining OOD datasets are imperative, not just procedural. Neglecting this essential aspect can render models susceptible to misinterpretation, diminished accuracy, and compromised robustness when faced with unanticipated data. Thus, Clean OOD datasets are crucial in equipping AI models to navigate and adapt to diverse and dynamic real-world contexts adeptly.

\subsection{Dataset Optimization for Enhanced OOD Detection}

Our study has concentrated on refining five prominent open-source datasets to enhance the fairness and accuracy of Out-of-Distribution (OOD) detection evaluation. This refinement process entailed rigorous image verification within each dataset, ensuring alignment with the corresponding synsets identified in our initial analysis. Our dataset selection includes  Places365\cite{zhou2017places}, Texture\cite{cimpoi2014describing}, iNaturalist\cite{van2018inaturalist}, SUN subset\cite{xiao2010sun}, and ImageNet-O\cite{hendrycks2021natural}.. The primary goal of this integration is to improve the accuracy of class classification within our dataset evaluation, ensuring that it genuinely represents the true nature of the images.

Our method categorises images from these datasets into 1,000 categories recognized by the ImageNet-1K classification model. This meticulous categorisation process aims to assign each image accurately to its correct type despite challenges such as occlusions, distracting elements, and multiple types within many images. We employ a multi-user independent classification system to ensure a broad spectrum of representation and precise labelling. An image is classified as in-distribution (ID) only if it garners substantial majority consensus among reviewers; in the absence of such consensus, it is considered an OOD sample. This approach mitigates the risk of low-confidence classifications.

Additionally, we provide comprehensive documentation of the ID data category labels in the OOD dataset and utilize cosine similarity metrics for visual similarity analysis. Several methodologies are implemented to refine the quality of annotations further. Annotators uncertain about an image's category can label it as "Uncategorized," signaling the need for further review. Each image undergoes evaluation by at least five independent annotators, with consistent results guiding its final classification. This process includes multiple rounds of filtering and regular quality checks to uphold high annotation standards. For particularly challenging images, especially in ImageNet-O\cite{hendrycks2021natural}, we seek additional reviewer input to more accurately capture category complexity.

\begin{figure*}[!t]
\centering
\includegraphics[width=7in]{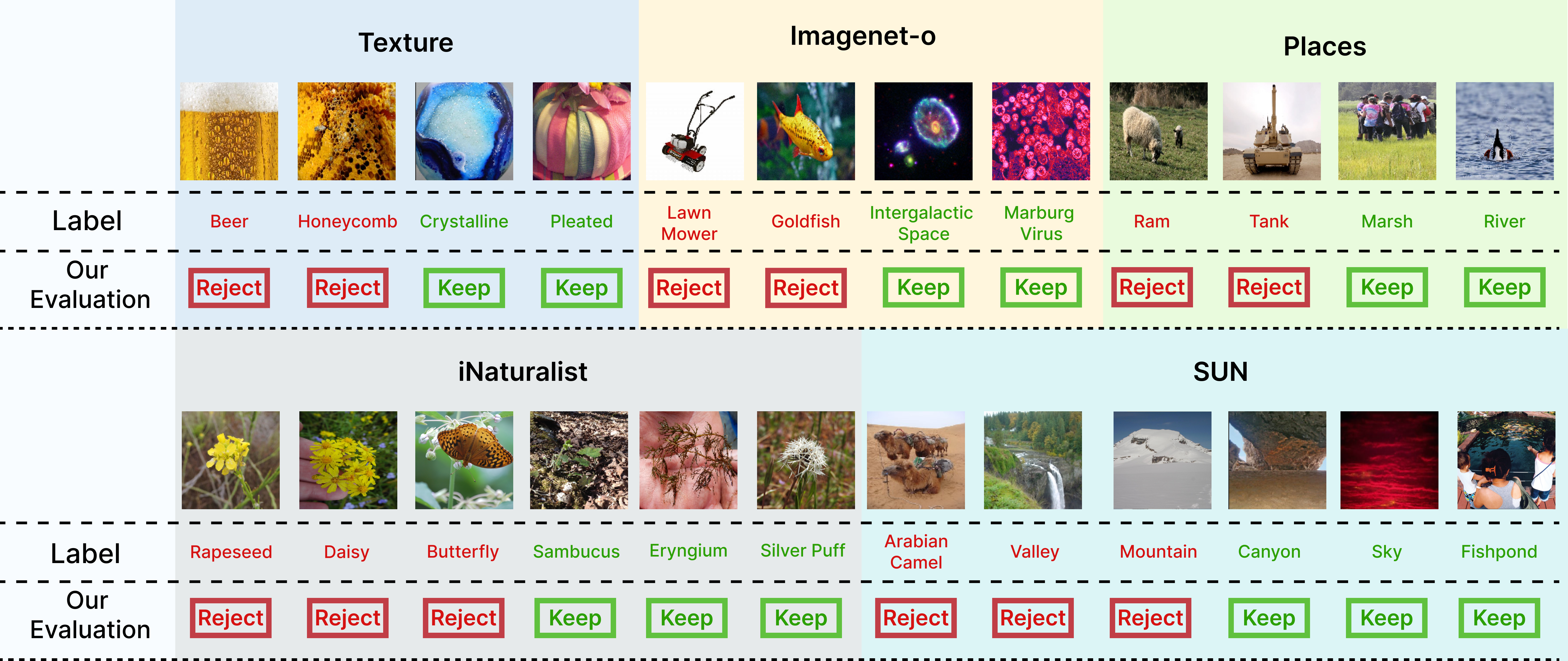}
\caption{Categorization of data samples within various OOD datasets. This figure demonstrates the classification results where each image is labeled as either ID or OOD across different datasets. Red boxes indicate images incorrectly labeled as ID within OOD datasets (false negatives), and green boxes signify correctly identified OOD samples (true positives). The results highlight the challenge of distinguishing complex patterns in OOD detection tasks and the importance of accurate labeling for the optimization of OOD algorithms.}
\label{fig_2}
\end{figure*}

A crucial aspect of our methodology is the careful separation and elimination of ID data from the OOD dataset. As shown in Fig.  \ref{fig_2}, this meticulous classification process results in a dataset predominantly composed of authentic OOD samples, enhancing the validity and fairness of our image classification and OOD detection evaluations. After extensive optimization, our curated dataset significantly reduces noise, leading to more reliable OOD detection. This enhanced dataset represents a novel combination of deep feature extraction and semantic analysis in image classification tasks, ensuring an equitable and accurate evaluation of OOD detection models.

The OOD-R dataset, resulting from our meticulous curation and evaluation, forms the foundation of our evaluation methods and model improvements. This dataset, a balanced mix of ID and OOD data, has undergone rigorous and multifaceted evaluation to ensure its diversity, completeness, and effectiveness in enhancing OOD detection capabilities within neural network models.

\subsection{Activation Function Design for OOD Detection}

Leveraging the OOD-R dataset, we have utilized models like BiT\cite{kolesnikov2020big} and VGG\cite{ding2021repvgg}, capitalizing on their exceptional classification and feature extraction capabilities. Our evaluation paradigm integrates a suite of OOD scoring functions, including MSP\cite{hendrycks2016baseline}, MaxLogit\cite{hendrycks2019scaling}, Energy\cite{liu2020energy}, ReAct\cite{sun2021react}, ViM\cite{wang2022vim}, Residual, GradNorm, Mahalanobis\cite{lee2018simple}, and KL-Matching\cite{hendrycks2019scaling}. Utilizing the OOD-R dataset for evaluation contributes to a fair assessment of the model's adaptability and generalization capabilities in out-of-distribution (OOD) context.

Considering that out-of-distribution sample detection is a single-sample hypothesis testing task—where a single sample is evaluated to produce its OOD score—the specificity of individual samples could potentially diminish detection performance.
Therefore, we aim to mitigate the impact of specificity by computing the expectation of a single sample within a certain neighborhood. Specifically, we depart from using the vanilla ReLU activation function, which only computes the activation value for a single input. In this paper, we calculate the expected activation values within the input's neighborhood, as fomulated below:

\begin{equation}
     g(\boldsymbol{x}) = \mathbb{E}_{\epsilon\sim p_{\beta}}\left[\text{ReLU}(\boldsymbol{x}-\epsilon)\right],
\label{neweq1}
\end{equation}
where $p_{\beta}(\epsilon)$ is implicitly defined.
Hence, the determination of whether the test sample is an OOD sample no longer relies solely on the activation of a single input but instead computes the average activation across the entire neighborhood. This approach contributes to more robust test results. In practice, for the sake of simplifying the computation process, we conduct the following derivation and simplification. The Eq. (\ref{neweq1}) can be reformulated in integral form as :
\begin{equation}
    g(\boldsymbol{x}) = \int_{-\infty}^{+\infty}p_\beta(\epsilon)\text{ReLU}(\boldsymbol{x}-\epsilon)d\epsilon.
\label{neweq2}
\end{equation}
With respect to $\boldsymbol{x}$, the differential of the Eq. \ref{neweq2} is:
\begin{equation}\begin{split}
     \frac{d}{dx} g(\boldsymbol{x}) = \int_{-\infty}^{+\infty}p_\beta(\epsilon)\Theta(\boldsymbol{x}-\epsilon)d\epsilon\ 
     = \int_{-\infty}^{\boldsymbol{x}}p_\beta(\epsilon)d\epsilon.
\end{split}
\label{neweq3}
\end{equation}
Here, we choose the $p_{\beta}(\epsilon)$ as:
\begin{equation}
    p_{\beta}(\epsilon) = \frac{\beta}{(e^{\beta\frac{\epsilon}{2}+e^{-\beta}\frac{\epsilon}{2}})^2}.
\label{neweq4}
\end{equation}
Then the Eq. (\ref{neweq3}) can be expressed as:
\begin{equation}\begin{split}
     \frac{d}{dx} g(\boldsymbol{x}) = \frac{1}{1 + \exp(-\beta x)}.
\end{split}
\label{neweq5}
\end{equation}
By integrating both sides of the equation, we can obtain:
\begin{equation}
    g(\boldsymbol{x}) = \frac{1}{\beta} \log(1 + \exp(\beta \boldsymbol{x})),
\label{neweq6}
\end{equation}
which is the calculation formula for the Softplus function. Combining Eq. (\ref{neweq6}) and (\ref{neweq1}), we have got an alternative to the expectation of activations within the neighborhood of input, which is convenient in practice.

As demonstrated in our detailed equations and analyses, the ActFun structure accentuates activation dynamics, fostering a more proficient neural network architecture in OOD detection. We exploit the intrinsic properties of the Softplus function to ensure smoother and more adaptable activation. This approach optimizes the model's response to diverse inputs, thereby improving its accuracy and reliability in environments with unpredictable data. In the Experiments section, we extensively discuss the impact of hyperparameters $\beta$ in Eq. (\ref{neweq6}).

\begin{figure*}[htbp]
\centering
\includegraphics[width=7in]{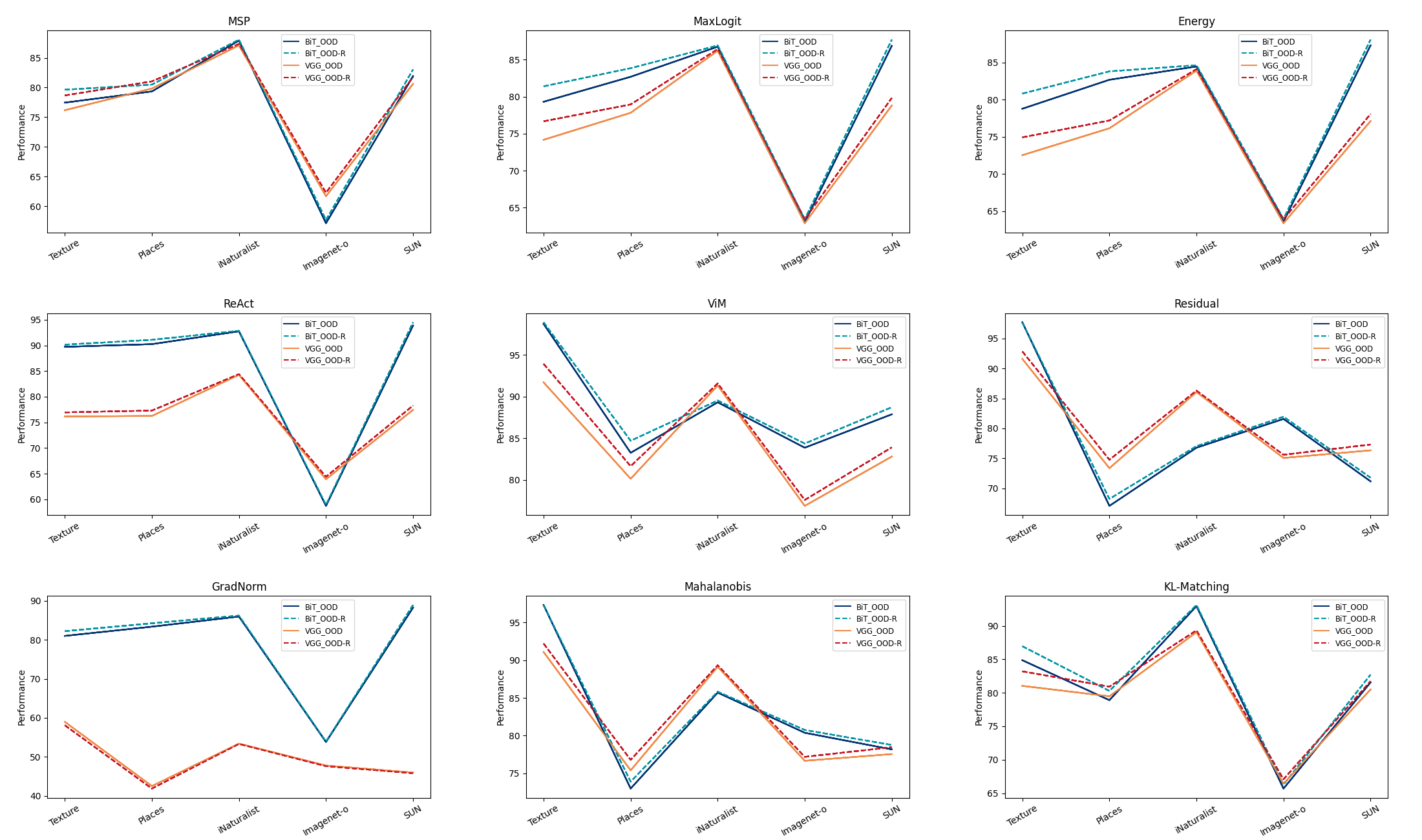}
\caption{Performance of various OOD detection algorithms across different datasets, before and after noise reduction. Each subplot represents a different detection method, with the solid lines indicating the detection performance on the original OOD datasets and the dashed lines showing performance on the noise-reduced datasets (OOD\-R). The blue and red lines correspond to the BiT\cite{kolesnikov2020big} and VGG\cite{ding2021repvgg} models, respectively. This analysis demonstrates the effects of noise reduction on the sensitivity and specificity of OOD detection methods, with varying degrees of impact observed across different methods and datasets.}
\label{fig_3}
\end{figure*}
\begin{table}[tbp]
    \fontsize{6.8pt}{10pt}\selectfont

	\begin{tabular}{cccccc}
		\toprule  % 顶部线
        \specialrule{0em}{2pt}{0pt}
            Dataset  & Texture & ImageNet-O & iNaturalist & Places365 & SUN subset\\
            \midrule  % 中部线   
            OOD  &  5640 & 2000 & 10000 & 10000 & 10000\\
            \textbf{OOD-R} & \textbf{5253} & \textbf{1933} & \textbf{9905} & \textbf{9449} & \textbf{9579} \\ 
		\bottomrule  % 底部线
	\end{tabular}
    \caption{The reduction in the number of samples in the corrected out-of-distribution dataset is used to refine the data set and reduce noise interference.}
    \label{table_1}
\end{table}
\section{Experiments}
In this section, we carefully evaluate the efficacy and applicability of our curated dataset, OOD-R, within the framework of comprehensive OOD detection tasks. Our evaluation strategy is multifaceted, designed to thoroughly scrutinize the robustness and validity of the dataset across various testing paradigms. Initially, our focus is on the well-established large-scale OOD detection benchmark\cite{huang2021mos} utilizing the ImageNet dataset. This stage, detailed in Section A, provides foundational insights into the performance characteristics of the OOD-R dataset, offering robust analysis within a recognized benchmarking environment. This ensures that our findings are comprehensive and comparable within the broader research community. Next, in Section B, we delve deeper into evaluating enhancements integrated into our approach. Here, we compare our methods with existing models on the BiT\cite{kolesnikov2020big} and VGG\cite{ding2021repvgg} networks to demonstrate the impact of these improvements on OOD detection capabilities. Finally, we find different results from the previous use of our proposed dataset under the influence of different hyperparameters $\beta$, see Section C. The results presented in this section highlight our method's advancements, contributing to the overall assessment of the OOD-R dataset's performance and applicability.

\subsection{Enhancing Datasets for Improved Data Quality Standards}
To elevate data quality standards and address the limitations imposed by noise, we introduce the open-source dataset group OOD-R, as shown in Table \ref{table_1}. This innovative dataset employs noise filtering technology to provide a sample repository with enhanced clarity and reliability. Our comprehensive evaluation, using models like BiT\cite{kolesnikov2020big} and VGG\cite{ding2021repvgg}, demonstrates significant improvements. We observed a 2.5\% increase in AUROC using MaxLogit\cite{hendrycks2019scaling} and a substantial 3.2\% reduction in FPR95 with ViM\cite{wang2022vim}. Fig. \ref{fig_3} graphically represents these findings, emphasizing the crucial role of datasets in improving assessment accuracy and reliability. Our dataset's unique low noise characteristic, extensively discussed in the Results section, provides context for understanding these experimental results and underscores its contribution to enhancing the accuracy and credibility of OOD detection methods.
\begin{table*}[htbp]

\setlength\tabcolsep{1pt} 
	\centering
    \fontsize{6.8pt}{10pt}\selectfont

	\begin{tabular}{ccccccccccccc}
 
		\toprule  % 顶部线
	%	  \multicolumn{2}{c|}{Model} 
        \multirow{2}{*}{Method}&\multicolumn{2}{c}{Texture}&\multicolumn{2}{c}{Places}&\multicolumn{2}{c}{iNaturalist}&\multicolumn{2}{c}{Imagenet-o}&\multicolumn{2}{c}{SUN}&\multicolumn{2}{c}{Average} \\
            & AUROC$\uparrow$&FPR95$\downarrow$&AUROC$\uparrow$&FPR95$\downarrow$&AUROC$\uparrow$&FPR95$\downarrow$&AUROC$\uparrow$&FPR95$\downarrow$&AUROC$\uparrow$&FPR95$\downarrow$&AUROC$\uparrow$&FPR95$\downarrow$\\
        \hline
        \specialrule{0em}{2pt}{0pt}
            GradNorm &     82.20 & 57.15 & 84.22 & 59.89 & 86.19 & 58.22 & 53.92 & 91.77 & 88.90 & 43.81 & 79.09 & 62.17 \\
            GradNorm\_ActFun & {\textbf{85.12(+2.92)}} & {\textbf{50.64(-6.51)}} & {\textbf{87.51(+3.29)}} & {\textbf{50.5(-9.39)}}  & {\textbf{90.74(+4.55)}} & {\textbf{42.48(-15.74)}} & {\textbf{59.39(+5.47)}} & {\textbf{88.72(-3.05)}} & {\textbf{92.18(+3.28)}} & {\textbf{33.78(-10.03)}} & {\textbf{82.99(+3.90)}} & {\textbf{53.23(-8.94)}}  \\    
        \hline
            ReAct  &     90.15 & 44.53 & 91.08 & 46.64 & 92.85 & 38.56 & 58.91 & 89.24 & 94.49 & 29.55 & 85.50 & 49.70 \\
            ReAct\_ActFun  & {\textbf{93.23(+3.08)}} & {\textbf{33.87(-10.66)}}& {\textbf{91.62(+0.54)}} & {\textbf{42.70(-3.94)}} & {\textbf{95.11(+2.26)}} & {\textbf{26.13(-12.43)}} & {\textbf{63.23(+4.32)}} & {\textbf{85.46(-3.78)}} & {\textbf{95.06(+0.57)}} & {\textbf{26.08(-3.47)}}  & {\textbf{87.65(+2.15)}} & {\textbf{42.85(-6.85)}}  \\      
        \hline
            Mahalanobis &  97.31 & 14.32 & 73.88 & 81.84 & 85.82 & 64.79 & 80.74 & 69.63 & 78.75 & 72.78 & 83.30 & 60.67 \\
            Mahalanobis\_ActFun & {\textbf{98.31(+1.00)}} & {\textbf{8.32(-6.00)}}  & {\textbf{74.23(+0.35)}} & {\textbf{80.07(-1.77)}} & {\textbf{87.83(+2.01)}} & {\textbf{59.72(-5.07)}}  & {\textbf{82.74(+2.00)}} & {\textbf{63.99(-5.64)}} & {\textbf{80.20(+1.45)}} & {\textbf{68.78(-4.00)}}  & {\textbf{84.66(+1.36)}} & {\textbf{56.18(-4.49)}} \\               
        \hline
            Energy   &     80.83 & 74.41 & 83.82 & 72.02 & 84.65 & 74.77 & 63.97 & 96.22 & 88.09 & 59.69 & 80.27 & 75.42 \\
            Energy\_ActFun   & {\textbf{82.69(+1.86)}} & {\textbf{68.34(-6.07)}} & 83.60    & {\textbf{71.69(-0.33)}} & {\textbf{85.74(+1.09)}} & {\textbf{70.16(-4.61)}}  & {\textbf{66.12(+2.15)}} & {\textbf{95.45(-0.77)}} & {\textbf{88.21(+0.12)}} & {\textbf{58.94(-0.75)}}  & {\textbf{81.27(+1.00)}} & {\textbf{72.92(-3.50)}}  \\    
        \hline
            MaxLogit &     81.40 & 74.05 & 83.85 & 73.16 & 86.93 & 70.32 & 63.42 & 96.84 & 87.71 & 62.39 & 80.66 & 75.35 \\
            MaxLogit\_ActFun & {\textbf{82.64(+1.24)}} & {\textbf{69.88(-4.17)}} & 83.73   & {\textbf{72.21(-0.95)}} & {\textbf{88.04(+1.11)}} & {\textbf{64.26(-6.06)}}  & {\textbf{65.31(+1.89)}} & {\textbf{95.91(-0.93)}} & {\textbf{87.87(+0.16)}} & {\textbf{61.49(-0.90)}}  & {\textbf{81.52(+0.86)}} & {\textbf{72.75(-2.60)}}  \\        
        \hline        
            MSP  & 79.63 & 77.42 & 80.49 & 78.02 & 88.07 & 64.38 & 57.67 & 96.90 & 83.04 & 70.92 & 77.78 & 77.53\\
            MSP\_ActFun  & 79.56    & {\textbf{75.56(-1.86)}} & {\textbf{80.56(+0.07)}} & {\textbf{77.25(-0.77)}} & {\textbf{88.55(+0.48)}} & {\textbf{61.34(-3.04)}}  & 57.49      & 96.90    & {\textbf{83.09(+0.05)}} & {\textbf{69.89(-1.03)}}  & {\textbf{77.85(+0.07)}} & {\textbf{76.19(-1.34)}}  \\                                            
		\bottomrule  % 底部线
	\end{tabular}
 \captionsetup{font=small}
 \caption{The upper table presents the performance metrics of the BiT model's OOD detection capabilities, following the substitution of the traditional ReLU activation function with ActFun. The metrics reported include AUROC and FPR95. Each method's performance is evaluated to ascertain the impact of the ActFun modification on the model's OOD detection efficiency.Among them, “$\uparrow$" represents that the larger the value, the better, and “$\downarrow$” represents that the smaller the value, the better. Our method is written as method\_ActFun, the best-performing items are shown in bold, and the increase or decrease numbers are in parentheses.}
 \label{table_2}
\end{table*}
\begin{table*}[htbp]
\setlength\tabcolsep{0.8pt} 
	\centering
    \fontsize{6.6pt}{10pt}\selectfont
	\begin{tabular}{ccccccccccccc}
    
		\toprule  % 顶部线
	%	  \multicolumn{2}{c|}{Model} 
        \multirow{2}{*}{Method}&\multicolumn{2}{c}{Texture}&\multicolumn{2}{c}{Places}&\multicolumn{2}{c}{iNaturalist}&\multicolumn{2}{c}{Imagenet-o}&\multicolumn{2}{c}{SUN}&\multicolumn{2}{c}{Average} \\
            & AUROC$\uparrow$&FPR95$\downarrow$&AUROC$\uparrow$&FPR95$\downarrow$&AUROC$\uparrow$&FPR95$\downarrow$&AUROC$\uparrow$&FPR95$\downarrow$&AUROC$\uparrow$&FPR95$\downarrow$&AUROC$\uparrow$&FPR95$\downarrow$\\
        \hline
        \specialrule{0em}{2pt}{0pt}
            GradNorm  & 58.10 & 91.28 & 41.89 & 98.88 & 53.33 & 98.20 & 47.63 & 95.65 & 45.83 & 98.20 & 49.35 & 96.44  \\
            GradNorm\_ActFun  & {\textbf{70.46(+12.36)}} & {\textbf{87.40(-3.88)}} & {\textbf{67.29(+25.40)}} & {\textbf{91.67(-7.21)}} & {\textbf{76.56(+23.23)}}& {\textbf{90.12(-8.08)}} & {\textbf{50.66(+3.03)}} & {\textbf{95.29(-0.36)}} & {\textbf{73.86(+28.03)}} & {\textbf{87.91(-10.29)}} & {\textbf{67.77(+18.42)}} & {\textbf{90.48(-5.96)}} \\
        \hline
            Energy  & 74.94 & 82.73 & 77.21 & 83.37 & 84.11 & 75.40 & 63.91 & 87.84 & 78.08 & 83.63 & 75.65 & 82.60  \\
            Energy\_ActFun  & {\textbf{79.11(+4.17)}} & {\textbf{69.07(-13.66)}}& {\textbf{84.42 (+7.21)}}& {\textbf{63.93(-19.44)}} & {\textbf{90.12(+6.01)}} & {\textbf{50.22(-25.18)}} & 63.82 &         {\textbf{85.36(-2.48)}} & {\textbf{86.17(+8.09)}} & {\textbf{59.80(-23.83)}} & {\textbf{80.73(+5.08)}} & {\textbf{65.67(-16.93)}} \\
        \hline
            ReAct  & 76.95 & 82.12 & 77.30 & 83.13 & 84.40 & 74.81 & 64.41 & 87.74 & 78.27 & 83.03 & 76.27 & 82.17  \\
            ReAct\_ActFun  & {\textbf{81.09(+4.14)}} & {\textbf{68.15(-13.97)}} & {\textbf{84.18(+6.88)}} & {\textbf{63.75(-19.38)}} &         {\textbf{90.23(+5.83)}} & {\textbf{49.39(-25.42)}}    &         64.27 &         {\textbf{85.31(-2.31)}} &         {\textbf{85.97(+7.70)}} & {\textbf{59.46(-23.57)}} &         {\textbf{81.15(+4.88)}} & {\textbf{65.21(-16.96)}} \\
        \hline            
            MaxLogit  & 76.67 & 76.66 & 78.95 & 77.36 & 86.43 & 61.91 & 63.43 & 89.71 & 79.83 & 77.00 & 77.06 & 76.53  \\
            MaxLogit\_ActFun  & {\textbf{79.98(+3.31)}} & {\textbf{66.70(-9.96)}} & {\textbf{84.81(+5.86)}} & {\textbf{62.48(-14.88)}} & {\textbf{91.47(+5.04)}} & {\textbf{41.53(-20.38)}} & 63.43 &         {\textbf{88.00(-1.71)}} & {\textbf{86.37(+6.54)}} & {\textbf{59.09(-17.91)}} & {\textbf{81.21(+4.15)}} & {\textbf{63.56(-12.97)}} \\
        \hline
            MSP  & 78.66 & 72.64 & 81.04 & 74.03 & 87.34 & 54.65 & 62.27 & 91.41 & 81.76 & 72.86 & 78.22 & 73.12      \\
            MSP\_ActFun  & {\textbf{80.41(+1.75)}} & {\textbf{67.33(-5.31)}} & {\textbf{84.03(+2.99)}} & {\textbf{64.57(-9.46)}} &  {\textbf{91.09(+3.75)}} & {\textbf{41.15(-13.50)}} &  {\textbf{62.82(+0.55)}} & {\textbf{89.91(-1.50)}} & {\textbf{85.30(+3.54)}} & {\textbf{62.30(-10.56)}} & {\textbf{80.73(+2.51)}} & {\textbf{65.05(-8.07)}}\\
        \hline
            KL-Matching & 83.21 & 61.55 & 80.92 & 74.78 & 89.33 & 41.95 & 67.08 & 84.69 & 81.73 & 74.03 & 80.45 & 67.40  \\
            KL-Matching\_ActFun & {\textbf{83.37(+0.16)}} & 61.96 &         {\textbf{81.00(+0.08)}} & 75.20 &         {\textbf{89.51(+0.18)}} & {\textbf{41.29(-0.66)}} & {\textbf{67.10(0.02)}} & {\textbf{84.27(-0.42)}} & {\textbf{81.80(+0.07)}} & 74.11 &         {\textbf{80.56(+0.11)}} & {\textbf{67.37(-0.03)}} \\  

		\bottomrule  % 底部线
	\end{tabular}
  \captionsetup{font=small}
 \caption{The lower table details the evaluation of the VGG model's OOD detection after integrating the ActFun function in place of ReLU. Similar to the BiT model, this table reports the AUROC and FPR95 metrics, offering a comparative view of the performance across the same diverse datasets, which enables a direct assessment of how the Softplus function influences the VGG model's ability to discriminate between in-distribution and OOD samples. The table also summarizes the average performance across all datasets, providing a holistic view of the effectiveness of the ActFun adaptation.}
 \label{table_3}
\end{table*}
In dataset optimization, the observed impact on OOD detection algorithms is intricately linked to the unique attributes each algorithm leverages. Algorithms like MaxLogit\cite{hendrycks2019scaling}, Energy\cite{liu2020energy}, Mahalanobis\cite{lee2018simple}, and KL-Matching\cite{hendrycks2019scaling} show significant performance variability due to their reliance on model confidence and data distribution assumptions. MaxLogit\cite{hendrycks2019scaling} and Energy\cite{liu2020energy} are heavily influenced by model prediction confidence; thus, optimizations altering decision boundaries or confidence scores can markedly impact their effectiveness. The Mahalanobis\cite{lee2018simple} method presumes data points to cluster around a central mean in feature space, and reductions in dataset size can alter the mean and covariance estimates, profoundly affecting performance through changes in distance calculations. Similarly, KL-Matching\cite{hendrycks2019scaling} evaluates the divergence between the predicted probabilities of in-distribution and OOD samples, with dataset optimizations potentially leading to a more uniform distribution that heightens the sensitivity of KL divergence to the remaining data points, substantially influencing algorithm performance.

Conversely, MSP\cite{hendrycks2016baseline}, ReAct\cite{sun2021react}, and GradNorm exhibit stability across datasets despite size reductions. MSP\cite{hendrycks2016baseline}'s dependence on the maximum softmax probability output means that dataset downsizing doesn't necessarily disrupt the distribution of these probabilities, thus maintaining stable performance. ReAct\cite{sun2021react}'s method, which adjusts network activations to mitigate adversarial perturbations, is less dependent on exact data distributions, relying more on network activation patterns, making it inherently robust to dataset size changes. Similarly, GradNorm's focus on the gradient norm as an OOD signal ties less to data distribution and more to the model's response, leaving its performance relatively unscathed by dataset size reductions.
Overall, the differential impact of dataset optimization on OOD detection methods stems from each algorithm's interaction with the dataset's statistical properties and model confidence measures. Algorithms that utilize detailed statistical analysis of the dataset exhibit a higher sensitivity to its changes. In contrast, those employing broader data features or model dynamics showcase consistent performance resilience in the face of dataset variability. This understanding is critical for tailoring dataset optimization strategies to each OOD detection algorithm's specific requirements and strengths.
\begin{figure*}[htbp]
\centering
\includegraphics[width=7in]{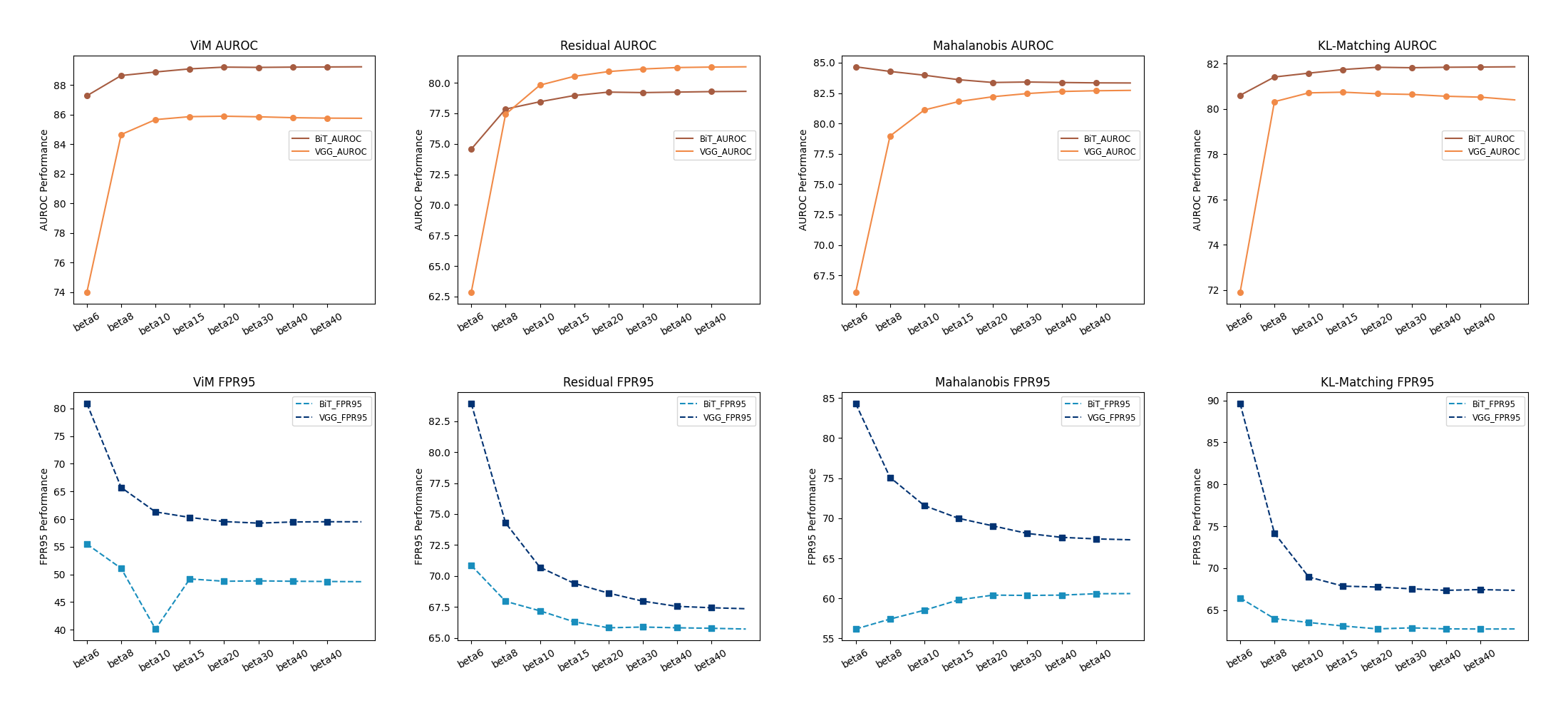}
\caption{ Comparative performance of OOD detection methods with varying $\beta$ values for the Softplus activation function. ViM\cite{wang2022vim}, KL-Matching\cite{hendrycks2019scaling}, Residual and Mahalanobis\cite{lee2018simple} demonstrate how an increase in beta affects assay sensitivity and specificity. The ViM\cite{wang2022vim} and KL-Matching\cite{hendrycks2019scaling} methods show improved or stable detection rates as $\beta$ increases, whereas the Residual and Mahalanobis\cite{lee2018simple} methods exhibit increased false positives, indicating a sensitivity to the activation function's smoothness. The results highlight the critical role of $\beta$ in balancing model sensitivity and robustness for OOD detection.}
\label{fig_4}
\end{figure*}
\subsection{Comparison of evaluation methods}
In computer vision, the capability of models to effectively process data from diverse sources is crucial, underscoring the importance of evaluating their robustness and generalization abilities on unknown data. To this end, we have conducted a comprehensive assessment using large-scale OOD detection benchmarks, incorporating a variety of OOD test datasets. These include subsets from Places365\cite{zhou2017places}, Texture\cite{cimpoi2014describing}, iNaturalist\cite{van2018inaturalist}, SUN, and ImageNet-O\cite{hendrycks2021natural}. The results, as presented in Tables \ref{table_2} and \ref{table_3}, offer a detailed analysis of the model's performance across different scenarios, aiming to enhance its robustness and generalization capabilities in real-world applications.

Our study critically examines the BiT\cite{kolesnikov2020big} and VGG\cite{ding2021repvgg} models in OOD detection tasks, highlighting the transition from traditional ReLU activation to Softplus. This modification is intended to harness Softplus’s gradient-preserving and differentiable attributes, thereby increasing the model's sensitivity to OOD instances. As delineated in two tables, we observed that methods like GradNorm, ReAct\cite{sun2021react}, and MaxLogit\cite{hendrycks2019scaling} significantly benefit from Softplus's consistent gradient flow and smooth transitional activations. This adaptation enhances their ability to discriminate between in-distribution and OOD data. Similarly, the Energy\cite{liu2020energy} method and MSP\cite{hendrycks2016baseline} also show improvements, attributable to the expanded logit range and more informative softmax probabilities, resulting in more precise OOD detection.

The application of Softplus in the VGG model corroborates these findings. These results highlight the complexity of selecting appropriate activation functions for OOD detection, emphasizing that enhancements beneficial for some methods may adversely affect others. By replacing traditional ReLU with Softplus, ActFun aims to capitalize on the latter’s consistent gradient and smooth activation transitions. This integration significantly enhances performance metrics such as FPR95 and AUROC. Specifically, GradNorm demonstrates marked improvements, evidenced by better AUROC scores and reduced FPR95, indicating enhanced accuracy in distinguishing subtle differences between in-distribution and OOD data. The ReAct\cite{sun2021react} method also exhibits improved performance, especially in datasets like iNaturalist\cite{van2018inaturalist} and SUN\cite{xiao2010sun}, benefiting from the refined control over network activations enabled by Softplus. These findings validate that ActFun can effectively augment OOD detection methods, leveraging the differentiable nature of Softplus and its ability to preserve gradient information, which is crucial for gradient-based methods like GradNorm and activation adjustment techniques such as ReAct\cite{sun2021react}.

\subsection{Impact of hyperparameter $\beta$ on results}

The experimental data in Fig. \ref{fig_4} elucidate the effects of the Softplus activation function's hyperparameter $\beta$ on OOD detection methods. The ViM\cite{wang2022vim} method, which employs a probabilistic model for uncertainty, shows improved or stable AUROC values and a decline in FPR95 as $\beta$ increases. This trend indicates that a milder slope in the activation function aids in better representing the probabilistic aspects of the data, leading to more accurate uncertainty estimation, a critical factor in OOD detection.

KL-Matching\cite{hendrycks2019scaling} relies on the Kullback-Leibler divergence for measuring the discrepancy between ID and OOD probability distributions. The maintenance of AUROC across different $\beta$ values suggests KL-Matching's robustness against variations in activation smoothness. However, a decrease in FPR95 with higher $\beta$ values implies that a more distinct activation response enhances the method's ability to differentiate between data distributions, thereby improving OOD reject rates.

The Residual Method employs skip connections to maintain gradient flow and achieves high AUROC scores, signifying effective OOD sample identification. Nonetheless, the increase in FPR95 observed with larger $\beta$ values points to potential over-smoothing within the feature space, possibly weakening the distinctive features that Residual connections aim to preserve and leading to less clear ID-OOD separation at the decision boundary.

The Mahalanobis\cite{lee2018simple} method, noted for its effectiveness in high-dimensional spaces and based on a Gaussian distribution assumption of ID data, shows an increase in FPR95 with rising $\beta$ values. This sensitivity suggests that larger $\beta$ values, which more closely approximate ReLU, could disturb the Gaussian distribution assumption in feature space, compromising the clarity of distinctions necessary for OOD detection.

In summary, the ViM\cite{wang2022vim} and KL-Matching\cite{hendrycks2019scaling} methods appear to capitalize on both increased and decreased smoothness afforded by the Softplus function. In contrast, the Residual and Mahalanobis\cite{lee2018simple} methods exhibit a nuanced response to $\beta$, where the former suffers from increased false positives at higher $\beta$ values, and the latter shows diminished performance, possibly due to misaligned Gaussian distribution assumptions. This complex interplay between $\beta$ and OOD detection efficacy accentuates the importance of method-specific hyperparameter tuning. It is essential to comprehend the interaction between each algorithm's core mechanics and the activation function to fine-tune performance, particularly when modifying key model components such as activation functions.

\section{Conclusion}
Our work introduces the open-source dataset OOD-R and the novel method ActFun, marking significant strides in enhancing OOD detection in neural networks. With its noise filtering technologies, OOD-R boasts low-noise characteristics that achieve up to a 2.5\% improvement in accuracy and a minimum 3.2\% reduction in false positives in a given network. It facilitates the extraction of cleaner, more reliable samples. This results in a more accurate and trustworthy evaluation. Empirical evidence from rigorous experiments and analyses across various domains and tasks demonstrates notable performance improvements from our approach. Furthermore, ActFun represents a blend of innovative technical adjustments and deep theoretical insights, recalibrating the neural network's input response. This brings significant improvements to the OOD-R dataset, increasing the performance of the GradNorm method by 18.42\% and reducing the false positive rate of the Energy method by 16.93\%.  It effectively reduces the influence of hidden units on OOD output and enhances data separability, leading to improved results in specific networks. Furthermore, our research elucidates the intricate interplay between the hyperparameter $\beta$ and the efficacy of various OOD detection algorithms. We underscore the imperative for meticulous hyperparameter tuning and an in-depth understanding of each algorithm's underlying principles.  ActFun's theoretical underpinnings provide valuable insights into neural network mechanisms in OOD scenarios, making it a practical and adaptable method for image and multi-class classification applications. Our approach contributes to the current understanding of OOD detection within neural networks and opens avenues for future research. We anticipate extending these methods beyond image classification to deepen and enrich the exploration of OOD detection mechanisms across various neural network applications.

\bibliographystyle{IEEEtran}
\bibliography{References}
 
\vspace{11pt}

\vfill

\end{document}